\documentclass[letterpaper, 10 pt, journal, twoside]{IEEEtran}  %

\IEEEoverridecommandlockouts                              %

\usepackage[caption=false,font=footnotesize]{subfig}
\usepackage{graphics} %
\usepackage{epsfig} %
\usepackage{amsmath} %
\usepackage{amsfonts} %
\usepackage{amssymb}  %
\usepackage{mathtools}
\usepackage{bm}
\usepackage{bbm}
\usepackage[hidelinks]{hyperref}
\usepackage{siunitx}
\usepackage{import}
\usepackage{booktabs}
\usepackage{placeins}
\usepackage{tensor}
\usepackage{tikz}
\usepackage{tikz_utils}  %
\usepackage[subpreambles=false]{standalone}
\usepackage{bigdelim}
\usepackage{array}
\usepackage{gensymb}
\usepackage{makecell}
\usepackage{threeparttable}
\usepackage{colortbl}
\usepackage{boldline}
\usepackage{pgfplots}

\definecolor{lightblue}{rgb}{0.678,0.847,0.902}

\usetikzlibrary{calc}
\usetikzlibrary{positioning}
\usetikzlibrary{patterns}
\usetikzlibrary{decorations,backgrounds,shapes,intersections}
\usepgfplotslibrary{units}

\pgfplotsset{compat=1.16} 
\pgfplotsset{every tick label/.append style={font=\footnotesize}}
\pgfplotsset{every axis label/.append style={font=\footnotesize}}
\pgfplotsset{every legend /.append style={font=\footnotesize}}

\DeclareMathOperator*{\argmax}{arg\,max}

\newcommand{\hl}[1]{}

\newcommand{\hln}[1]{#1}

\newcolumntype{B}{>{\columncolor{green!25}}r}

\newcommand\Tstrut{\rule{0pt}{3ex}}       %
\newcommand\Bstrut{\rule[-1.4ex]{0pt}{0pt}} %
\newcommand{\TBstrut}{\Tstrut\Bstrut} %

\newcolumntype{L}[1]{>{\raggedright\let\newline\\\arraybackslash\hspace{0pt}}m{#1}}
\newcolumntype{C}[1]{>{\centering\let\newline\\\arraybackslash\hspace{0pt}}m{#1}}
\newcolumntype{R}[1]{>{\raggedleft\let\newline\\\arraybackslash\hspace{0pt}}m{#1}}

\title{SDFEst: Categorical Pose and Shape Estimation of Objects from RGB-D using Signed Distance Fields}

\author{Leonard Bruns$^{1}$ and Patric Jensfelt$^{1}$%
\thanks{Manuscript received: February, 24, 2022; Revised May, 20, 2022; Accepted June, 20, 2022.} %
\thanks{This paper was recommended for publication by Editor Cesar Cadena upon evaluation of the Associate Editor and Reviewers' comments.} %
\thanks{$^{1}$The authors are with the Division of Robotics, Perception and Learning (RPL), KTH Royal Institute of Technology, Stockholm, Sweden {\tt\footnotesize\{leonardb,patric\}@kth.se}.}
}%

\begin{document}
\maketitle%
\begin{abstract}
    Rich geometric understanding of the world is an important component of many robotic applications such as planning and manipulation. In this paper, we present a modular pipeline for pose and shape estimation of objects from RGB-D images given their category. The core of our method is a generative shape model, which we integrate with a novel initialization network and a differentiable renderer to enable 6D pose and shape estimation from a single or multiple views. We investigate the use of discretized signed distance fields as an efficient shape representation for fast analysis-by-synthesis optimization. Our modular framework enables multi-view optimization and extensibility. We demonstrate the benefits of our approach over state-of-the-art methods in several experiments on both synthetic and real data. We open-source our approach at \url{https://github.com/roym899/sdfest}.
\end{abstract}

\section{Introduction}

\IEEEPARstart{W}{e} investigate the problem of joint pose and shape estimation of objects from RGB-D data. Pose estimation of known objects \cite{hodavn2020bop} and shape modeling of aligned objects \cite{park2019deepsdf,mescheder2019occupancy} have made significant progress in recent years, but the joint task has received less attention so far \cite{chen2020category,tian2020shape}. Assuming knowledge of the full 3D model and pose of an object is common in various classic robotic algorithms, such as motion planning and grasp computation, but is not easy to achieve from partial sensor information. Furthermore, pose and shape estimation at a category level could be used in a mapping context to create an object-based world representation \cite{sucar2020nodeslam}. Such object-based representations could, for example, enable interaction with objects in virtual or augmented reality when only partial sensor information is available.

\begin{figure}[htb]
    \centering
    \includegraphics[width=\linewidth]{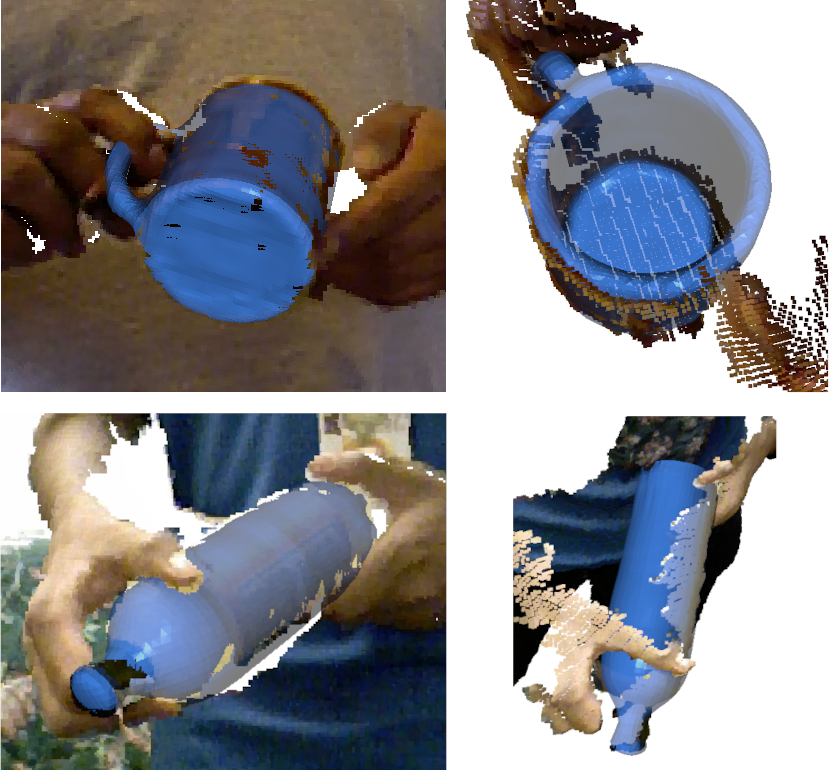}
    \caption{Single-view pose and shape estimation results for two objects obtained with SDFEst (left: front view, right: alternative view).}\label{fig:cool_pointcloud}
\end{figure}

Methods inferring image-based abstractions, such as classified bounding boxes and instance masks, have made remarkable progress in recent years due to the availability of large annotated datasets \cite{lin2014microsoft}. However, using such abstractions in a robotics context remains challenging. To bridge this gap from image-based abstractions to an actionable representation, we build on this progress by using a classified instance mask as the starting point to extract a cropped point set of the object of interest with the goal of subsequently estimating the full 3D shape and pose. %

Our work is inspired by the remarkable ability of humans to estimate the full shape of most objects from only a single view. This enables humans to grasp many objects without knowledge of the full shape or to plan more complicated tasks such as stacking. We hypothesize that this ability stems from integrating the prior knowledge accumulated from having seen many instances of an object category and the partial observation of the novel instance.

In this work, we model this intuition as a per-category generative object model. This generative model is trained to compress the shape variations of an object category in a low-dimensional representation. To then estimate pose and shape from partial sensor information, we propose to train a point-based network on synthetic data generated by the generative object model. We further refine this initial estimate using a differentiable renderer to better match the observations from a single or multiple views.

We present a modular architecture for single- and multi-view pose and shape estimation of objects from a known category. Our approach, SDFEst, only requires a collection of categorized and aligned meshes at training time and estimates the 6D pose and shape of an object at inference time (see Fig.~\ref{fig:cool_pointcloud}). Compared to purely discriminative approaches, \mbox{SDFEst} is modular in nature and allows optimizing the object's pose and shape from a single or multiple views by integrating a generative shape model with a differentiable renderer. Contrary to most existing approaches that use point sets, we use signed distance fields to represent shapes.

To summarize, our contributions are:
\begin{itemize}
    \item a novel extensible modular architecture for categorical pose and shape estimation from a single or multiple \mbox{RGB-D} images,
    \item a novel parametrization for multimodal orientation distributions, and
    \item an open-source implementation for pose and shape estimation using discretized signed distance fields (SDFs).
\end{itemize}
We compare our method to other related categorical pose and shape estimation methods and find that our method achieves state-of-the-art performance when poses are constrained and outperforms existing methods on unconstrained poses.

\section{Related Work}

We will summarize work of three related areas: RGB-based shape estimation, categorical pose and shape estimation, and optimizable shape and pose estimation.

\subsection{RGB-based Shape Estimation}

Several methods have been proposed to estimate the shape of an object from a single \cite{wu2016learning,fan2017point,groueix2018papier,wu2017marrnet,wang2018pixel2mesh,xu2019} or multiple RGB images \cite{choy20163d,xie2020pix2vox++}. See Han et al.~\cite{han2019image} for a recent survey on image-based 3D reconstruction.

In many cases, pose is not modeled explicitly and instead only the shape is predicted \cite{choy20163d, wu2016learning, groueix2018papier}. Although such an approach in principle can learn entangled representations of pose and shape, Zhu et al. \cite{zhu2017rethinking} showed that explicitly modeling pose and predict shapes in a canonical reference frame reduces the learning complexity significantly and further allows finetuning of the pose estimation on real-world silhouette annotations. 

Engelmann et al.~\cite{engelmann2021points} address the single RGB view multi-object case with a single-shot architecture. They frame the shape estimation problem as classification of the best matching shape. This allows them to decouple the shape estimation from the shape representation, but limits deformations to scalings of shapes in the database. Principled fusion of multiple such single-view predictions remains challenging.

In this work, we explicitly decouple pose, scale, and shape. We predict the shape in a canonical reference frame, which can be used to simplify downstream tasks such as grasp computation (e.g., grasping a mug from the top at the rim). %

\hl{
    \begin{itemize}
        \item Meshes (\cite{kanazawa2018learning})
        \item Occupancy (discretized \cite{choy20163d}, \cite{wu20153d}, \cite{girdhar2016learning}, continuous \cite{mescheder2019occupancy})
        \item Signed-distance field (coordinate-based \cite{park2019deepsdf})
        \item Distance field (discretized \cite{dai2017shape})
        \item Mem3D \cite{yang2021single} (read!)
    \end{itemize}
}

\hl{
    6/9D Pose Estimation
    Known cad models:
    \begin{itemize}
        \item Need to do check the literature for a few days
        \item CosyPose
        \item StablePose
        \item BOPBenchmark
    \end{itemize}
    Categorical:
    \begin{itemize}
        \item NOCS, 2019 \cite{wang2019normalized} (read)
        \item DualPoseNet, 2021, unpublished \cite{lin2021dualposenet}
        \item DONet, 2021, unpublished \cite{lin2021donet}
    \end{itemize}
}

\subsection{Categorical Pose and Shape Estimation}
While pose estimation of known objects has matured significantly \cite{hodavn2020bop}, pose estimation on a per-category level has only recently received more attention.

To estimate pose on a per-category level, \cite{wang2019normalized} proposed the normalized object coordinate space (NOCS). In this space, objects of one category are aligned in a unit cube. To estimate the object pose, the projected NOCS coordinates (also called NOCS map) are predicted from the RGB image. This NOCS map and the observed depth map can be considered as correspondences, which together with the Umeyama algorithm \cite{umeyama1991least} and RANSAC can be used to robustly estimate 6D pose and scale of the object. As part of their contribution, the authors also published the synthetic CAMERA dataset and the real-world REAL275 dataset. The latter being the most common dataset to evaluate categorical object pose estimation.

Building on these datasets proposed by \cite{wang2019normalized}, several methods were introduced to also address the categorical pose estimation problem. \cite{chen2020learning} proposed canonical shape space (CASS), which directly regresses a rotation matrix and translation vector from the observed point set and as a by-product also reconstructs the full point set. Shape prior deformation (SPD) \cite{tian2020shape} follows a similar idea but instead predicts deformations of a canonical point set for each category. To estimate the pose, they also use the NOCS. \cite{wang2021category} and \cite{chen2021sgpa} extend SPD with a recurrent architecture and a transformer architecture for better shape adaptation, respectively. All of these methods train on a mix of mixed-reality images from the CAMERA dataset and real images from the REAL dataset. In contrast, the recently proposed ASM-Net \cite{akizuki2021} also estimates pose and shape, but showed that competitive results can be obtained by training on synthetic renderings of meshes only. Similar to ASM-Net, we also only require a collection of meshes for training. %

\hln{Most methods (including ours) employ two-stage pipelines in which an object detection or instance segmentation module first detects bounding boxes or masks, which are later used to estimate the object's pose and shape. In contrast, Irshad et al.~\cite{irshad2022centersnap} proposed to use a single-shot architecture to detect objects and estimate their shape and pose jointly. While such an end-to-end approach might be easier to scale, data collection and data generation becomes more challenging compared to two-stage approaches, which can benefit from large-scale segmentation datasets.}

In prior work \cite{bruns2022evaluation}, we showed that existing methods do not generalize well to unconstrained orientations due to the constrained orientations present in the CAMERA and REAL datasets. In this work, we propose a method that can achieve competitive results without constraining the orientation and outperforms existing approaches when constraining orientations to those included in the training set.

\begin{figure*}[htb!]
    \centering
    \resizebox{\textwidth}{!}{\begin{tikzpicture}[scale=1]
    \coordinate (image_r) at (-5.8,1.4,0);%
    \coordinate (image_g) at (-5.6,1.4,0);%
    \coordinate (image_b) at (-5.4,1.4,0);%
    \coordinate (depth) at (-5.6,-1.4,0);%
    \coordinate (concat_depth) at (-0.2,0,0);%
    \coordinate (concat_mask) at (0,0,0);%
    \path[draw, dashed, thick, fill=gray!30!white, rounded corners, name path=latent] (3.3,-1.5) -- ++(4.6,0) -- ++(0, 0.8) -- ++(-3,0) --++(0,2.2) --++(-1.6,0) -- cycle;

    \draw[thin, contour=0.48\pgflinewidth, fill=red!30] (image_r) ++(0,0.6,0.9) -- ++(0,-1.2,0) -- ++(0.2,0,0) -- ++(0,1.2,0) -- cycle;%
    \draw[thin, contour=0.48\pgflinewidth, fill=red!30] (image_r) ++(0.2,0.6,0.9) -- ++(0,-1.2,0) -- ++(0,0,-1.8) -- ++(0,1.2,0) -- cycle;%
    \draw[thin, contour=0.48\pgflinewidth, fill=red!30] (image_r) ++(0,0.6,0.9) -- ++(0.2,0,0) -- ++(0,0,-1.8) -- ++(-0.2,0,0) -- cycle;%
    \draw[thin, contour=0.48\pgflinewidth, fill=green!30] (image_g) ++(0,0.6,0.9) -- ++(0,-1.2,0) -- ++(0.2,0,0) -- ++(0,1.2,0) -- cycle;%
    \draw[thin, contour=0.48\pgflinewidth, fill=green!30] (image_g) ++(0.2,0.6,0.9) -- ++(0,-1.2,0) -- ++(0,0,-1.8) -- ++(0,1.2,0) -- cycle;%
    \draw[thin, contour=0.48\pgflinewidth, fill=green!30] (image_g) ++(0,0.6,0.9) -- ++(0.2,0,0) -- ++(0,0,-1.8) -- ++(-0.2,0,0) -- cycle;%
    \draw[thin, contour=0.48\pgflinewidth, fill=blue!30] (image_b) ++(0,0.6,0.9) -- ++(0,-1.2,0) -- ++(0.2,0,0) -- ++(0,1.2,0) -- cycle;%
    \draw[thin, contour=0.48\pgflinewidth, fill=blue!30] (image_b) ++(0.2,0.6,0.9) -- ++(0,-1.2,0) -- ++(0,0,-1.8) -- ++(0,1.2,0) -- cycle;%
    \draw[thin, contour=0.48\pgflinewidth, fill=blue!30] (image_b) ++(0,0.6,0.9) -- ++(0.2,0,0) -- ++(0,0,-1.8) -- ++(-0.2,0,0) -- cycle;%
    \node[baseline=(X.base), anchor=east, text width=3mm] (image_i) at ($(image_g) + (0.0,-0.48,0.9)$) {$\mathbf{I}$};%
    \draw[thick] (image_b) ++(0.2,0,0) -- ++(0.8,0,0) node [draw=black, thick, align=center, anchor=west] (instance_segmentation) {Instance\\Segmentation};%
    \draw[thick] (instance_segmentation.east) -- ++(0.8,0,0) node[right] (mask_i) {$\mathcal{M}$};%
    \draw[thin, contour=0.48\pgflinewidth, fill=gray!30] (depth) ++(0,0.6,0.9) -- ++(0,-1.2,0) -- ++(0.2,0,0) -- ++(0,1.2,0) -- cycle;%
    \draw[thin, contour=0.48\pgflinewidth, fill=gray!30] (depth) ++(0.2,0.6,0.9) -- ++(0,-1.2,0) -- ++(0,0,-1.8) -- ++(0,1.2,0) -- cycle;%
    \draw[thin, contour=0.48\pgflinewidth, fill=gray!30] (depth) ++(0,0.6,0.9) -- ++(0.2,0,0) -- ++(0,0,-1.8) -- ++(-0.2,0,0) -- cycle;%
    \node[baseline=(X.base), anchor=east, text width=3mm] (image) at ($(depth) + (0.0,-0.48,0.9)$) {$\mathbf{D}$};%
    \draw[thick,<->] (depth) ++ (0.1,1.1,0) -- ($(image_g)+(0.1,-1.1,0)$) node [midway, right] {Registered};%
    \draw[thick] let
        \p1 = (mask_i.east),
        \p2 = (depth)
        in 
        (depth)  ++ (0.2,0,0) -- ($(\x1, \y2)-(0.2,0,0)$) to[out=0, in=-90] ($(concat_depth)+(0.1,-0.6,0)$);%
    \draw[thin, contour=0.48\pgflinewidth, fill=gray!30] (concat_depth) ++(0,0.6,0.9) -- ++(0,-1.2,0) -- ++(0.2,0,0) -- ++(0,1.2,0) -- cycle;%
    \draw[thin, contour=0.48\pgflinewidth, fill=gray!30] (concat_depth) ++(0.2,0.6,0.9) -- ++(0,-1.2,0) -- ++(0,0,-1.8) -- ++(0,1.2,0) -- cycle;%
    \draw[thin, contour=0.48\pgflinewidth, fill=gray!30] (concat_depth) ++(0,0.6,0.9) -- ++(0.2,0,0) -- ++(0,0,-1.8) -- ++(-0.2,0,0) -- cycle;%
    \draw[thin, contour=0.48\pgflinewidth, fill=gray!60] (concat_mask) ++(0,0.6,0.9) -- ++(0,-1.2,0) -- ++(0.2,0,0) -- ++(0,1.2,0) -- cycle;%
    \draw[thin, contour=0.48\pgflinewidth, fill=gray!60] (concat_mask) ++(0.2,0.6,0.9) -- ++(0,-1.2,0) -- ++(0,0,-1.8) -- ++(0,1.2,0) -- cycle;%
    \draw[thin, contour=0.48\pgflinewidth, fill=gray!60] (concat_mask) ++(0,0.6,0.9) -- ++(0.2,0,0) -- ++(0,0,-1.8) -- ++(-0.2,0,0) -- cycle;%
    \draw[thick] (mask_i.east) to[out=0, in=90] ($(concat_mask)+(0.1,0.6,0)$);%

    \draw[thick] (0.2,0,0) -- (1,0,0);%
    \draw[thick, contour=0.48\pgflinewidth, fill=blue!20] (1,0.6,0.6) -- (1,-0.6,0.6) -- (1.8,-0.3,0.3) -- (1.8,0.3,0.3) -- cycle;%
    \draw[thick, contour=0.48\pgflinewidth, fill=blue!20] (1.8,0.3,0.3) -- (1.8,-0.3,0.3) -- (1.8,-0.3,-0.3) -- (1.8,0.3,-0.3) -- cycle;%
    \draw[thick, contour=0.48\pgflinewidth, fill=blue!20] (1,0.6,0.6) -- (1.8,0.3,0.3) -- (1.8,0.3,-0.3) -- (1,0.6,-0.6) -- cycle;%

    \draw[thick] (1.8,0,0) -- (3.5,0,0); %
    \node[right, minimum width=1.2cm](z_obj) at ($(3.1,0.5,0)+(0.4,0.6,0)$) {$\mathbf{z}_\mathrm{shape}$};%
    \node[right, minimum width=4.2cm] (z_pose) at ($(3.1,-0.5,0) + (0.4,-0.6,0)$) {$\mathbf{z}_\mathrm{pose}=(\mathbf{z}_\mathrm{pos},\mathbf{z}_\mathrm{orientation})$} ;%
    \node[right, minimum width=1.2cm] (z_scale) at (3.5,0,0) {$z_\mathrm{scale}$};%
    \draw[thick] (1.8,0,0) to[out=0,in=180] (z_obj.west); %
    \draw[thick] (1.8,0,0) to[out=0,in=180] (z_pose.west);%

    \draw[thick] (z_obj.east) -- +(0.8,0) coordinate(sdf_vae_in);%
    \draw[thick, contour=0.48\pgflinewidth, fill=blue!20] (sdf_vae_in) ++(0,0.3,0.3) -- ++(0,-0.6,0) -- ++(0.8,-0.3,0.3) -- ++(0,1.2,0) -- cycle;%
    \draw[thick, contour=0.48\pgflinewidth, fill=blue!20] (sdf_vae_in) ++(0.8,0.6,0.6) -- ++(0,-1.2,0) -- ++(0,0,-1.2) -- ++(0,1.2,0) -- cycle;%
    \draw[thick, contour=0.48\pgflinewidth, fill=blue!20] (sdf_vae_in) ++(0,0.3,0.3) -- ++(0.8,0.3,0.3) -- ++(0,0,-1.2) -- ++(-0.8,-0.3,0.3) -- cycle;%
    \coordinate (sdf_vae_out) at ($(sdf_vae_in)+(0.8,0,0)$);%
    \draw[thick] (sdf_vae_out) -- ++(0.8,0,0);%
    \node[right] (sdf_volume) at ($(sdf_vae_out) +(0.8,0,0)$) {$\mathbf{\widetilde{V}}_\mathrm{SDF}$};%
    \path let 
        \p1 = (sdf_volume.east)
        in
        node [draw=black, thick, align=center, anchor=west] (diff_rend) at ($(\x1,0)+(0.8,0)$) {Differentiable\\Renderer} ;%
    \draw[thick] (z_scale.east) -- (diff_rend.west);%
    \path[draw, thick] let 
        \p1 = (sdf_volume.east), 
        \p2 = (z_pose) 
        in 
        (z_pose.east) -- (\x1, \y2) to[out=0,in=180] ($(diff_rend.west)+(0,-0.2)$);%
    \draw[thick] (sdf_volume.east) to[out=0,in=180] ($(diff_rend.west)+(0,0.2)$);%
    \coordinate (depth_out) at ($(diff_rend.east) + (0.8,0)$);%
    \draw[thick] (diff_rend.east) -- (depth_out);%
    \draw[thin, contour=0.48\pgflinewidth, fill=gray!40] (depth_out) ++(0,0.6,0.9) -- ++(0,-1.2,0) -- ++(0.2,0,0) -- ++(0,1.2,0) -- cycle;%
    \draw[thin, contour=0.48\pgflinewidth, fill=gray!40] (depth_out) ++(0.2,0.6,0.9) -- ++(0,-1.2,0) -- ++(0,0,-1.8) -- ++(0,1.2,0) -- cycle;%
    \draw[thin, contour=0.48\pgflinewidth, fill=gray!40] (depth_out) ++(0,0.6,0.9) -- ++(0.2,0,0) -- ++(0,0,-1.8) -- ++(-0.2,0,0) -- cycle;%
    \node[align=center, anchor=base] (depth_out_label) at ($(depth_out) + (0.1,-1.1,0.9)$) {$\mathbf{\widetilde{D}}$};%
    \coordinate (depth_comp) at ($(depth_out) + (1.4,0)$);%
    \draw[thin, contour=0.48\pgflinewidth, fill=gray!30] (depth_comp) ++(0,0.6,0.9) -- ++(0,-1.2,0) -- ++(0.2,0,0) -- ++(0,1.2,0) -- cycle;%
    \draw[thin, contour=0.48\pgflinewidth, fill=gray!30] (depth_comp) ++(0.2,0.6,0.9) -- ++(0,-1.2,0) -- ++(0,0,-1.8) -- ++(0,1.2,0) -- cycle;%
    \draw[thin, contour=0.48\pgflinewidth, fill=gray!30] (depth_comp) ++(0,0.6,0.9) -- ++(0.2,0,0) -- ++(0,0,-1.8) -- ++(-0.2,0,0) -- cycle;%
    \node[align=center, anchor=base] (depth_comp_label) at ($(depth_comp) + (0.1,-1.1,0.9)$) {$\mathbf{D}$};%
    \draw[thick, <->] ($(depth_out_label.base east) + (0,0.15)$) -- node [midway, below] (loss) {$L$} ($(depth_comp_label.base west) + (0,0.15)$);%
    \draw[thick, ->, rounded corners=0.2cm] (loss.south) -- ++ (0,-0.3) node (temp) {} -- (z_obj |- temp) -- (z_obj |- 0,-1.5);%
\end{tikzpicture}}
    \caption{Pipeline of SDFEst. We first apply instance segmentation and feed the masked point set into a PointNet-like network \cite{qi2017pointnet} which predicts the pose $\mathbf{z}_\mathrm{pose}$, scale $z_\mathrm{scale}$, and latent shape descriptor $\mathbf{z}_\mathrm{shape}$. The decoder decodes $\mathbf{z}_\mathrm{shape}$ into a discretized signed distance field (SDF)  $\mathbf{\widetilde{V}}_\mathrm{SDF}$ containing the full object shape in canonical pose. Given the pose, scale, and SDF, we render the depth map and iteratively optimize the latent representation by minimizing a loss $L$ between the rendered depth map $\widetilde{\mathbf{D}}$ and measured depth map $\mathbf{D}$.}\label{fig:architecture}
\end{figure*}
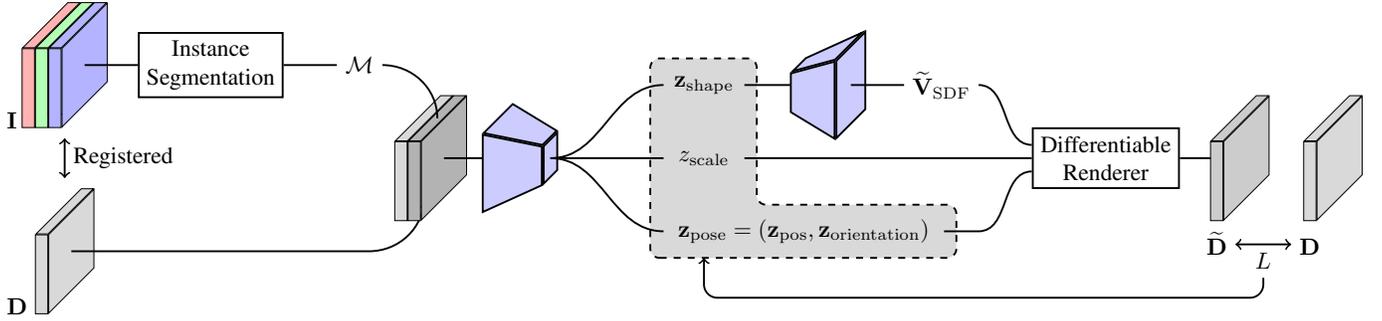

\subsection{Optimizable Pose and Shape Estimation}
Fewer works have investigated how to iteratively optimize the pose and shape given one or multiple observations. The approaches mentioned in the previous section are typically discriminative models, making it difficult to integrate additional information in a principled way. In contrast, the methods discussed in this section integrate a generative model into the estimation, which allows one to incorporate additional information by optimizing a latent representation, such that it matches one or multiple observations better.

FroDO \cite{runz2020frodo} is a framework for pose and shape estimation of objects from bounding box detections in multiple views. The approach uses keypoints tracked over multiple RGB frames to formulate a loss that refines the shape descriptor. Most notably, FroDO employs DeepSDF \cite{park2019deepsdf}, a continuous SDF representation, to represent the shape. MOLTR \cite{li2021moltr} is another RGB-based approach that also uses DeepSDF as the shape representation. MOLTR focuses on multi-object tracking, and instead of optimizing the shape descriptor through a loss, they fuse multiple single-view shape estimates through averaging. Compared to these works, we use discretized SDFs to represent shape and perform dense optimization on RGB-D data using a differentiable renderer.

Chen et al.~\cite{chen2020category} proposed an analysis-by-synthesis framework for pose and shape estimation of unknown objects. Based on a single RGB image, their approach samples and optimizes a large number of randomly sampled candidate poses to find the best matching candidate. Their approach does not allow direct extraction of the geometry; instead, novel views can be generated by their generative model. On the contrary, our method predicts an SDF, scale, and 6D pose, which can be used directly for geometric operations, such as collision checking or grasp planning.

\looseness=-1 In concurrent work, Deng et al.~\cite{deng2022icaps} introduced iCaps which, similar to our method, iteratively optimizes pose and shape from an initial estimate. Like FroDO and MOLTR, iCaps uses DeepSDF \cite{park2019deepsdf} to represent shapes. Instead, we investigate the use of discretized SDFs, which promises faster shape reconstruction and allow faster differentiable rendering, as shown in Section \ref{section:runtime}. For faster optimization, iCaps alternates between pose refinement and discriminative shape estimation given a pose. In contrast, our approach jointly optimizes pose and shape. Furthermore, due to iCaps' discriminative shape estimation, incorporating additional observations requires modifying the approach, while our gradient-based joint optimization directly supports incorporation of additional observations.

Our pipeline is inspired by NodeSLAM \cite{sucar2020nodeslam}, which employs a variational autoencoder (VAE) to model objects in a SLAM (simultaneous localization and mapping) framework. This VAE generates a discretized occupancy grid from a latent shape descriptor, which can be iteratively optimized using a probabilistic differentiable renderer. Orientation estimation in \cite{sucar2020nodeslam} is limited to a single axis by assuming objects to be upright on a table plane. Our work follows a similar idea to represent and optimize the object shape, but focuses on single-object, unconstrained, and possibly ambiguous object pose estimation.

\section{Problem Definition}

\looseness=-1 We study the problem of estimating 6D pose and shape at a per-category level. More formally, given an image $\mathbf{I}\in\mathbb{R}^{H\times W \times 3}$, depth map $\mathbf{D}\in\mathbb{R}^{H\times W}$, object mask $\mathcal{M}\subseteq \{(i,j)\in \mathbb{N}^2 \mid i\leq H, j \leq W\}$ and category $c\in \mathbb{N}$ we try to find estimates $\tensor[^{\mathrm{w}}]{\widetilde{\mathbf{T}}}{_{\mathrm{o}}}$ and $\widetilde{\mathcal{O}}$ of the true pose $\tensor[^{\mathrm{w}}]{\mathbf{T}}{_{\mathrm{o}}}$ and shape $\mathcal{O}$, respectively. \hln{We denote by $\mathrm{o}$ the object frame, by $\mathrm{w}$ the world frame, and by $\mathrm{c}$ the camera frame. We assume camera pose(s) $\tensor[^{\mathrm{w}}]{\mathbf{T}}{_{\mathrm{c}}}$ to be known.} Given the availability of depth maps, we consider metrically scaled shapes. %

\section{Method}\label{sec:method}

In this section, we will first describe the overall pipeline (Section \ref{sec:pipeline_overview}) and subsequently describe and motivate the design of the individual components (Section \ref{sec:shape_modeling} to \ref{sec:dr}). After introducing the components, we detail the inference step in Section \ref{sec:inference}. Finally, we provide further details of the training process in Section \ref{sec:training_details}.

\subsection{Pipeline Overview}\label{sec:pipeline_overview}

Fig.~\ref{fig:architecture} shows an overview of our proposed pipeline. The three main components are (from left to right) an initialization network (Section \ref{sec:init}), a generative shape model (Section \ref{sec:shape_modeling}), and a differentiable renderer (Section \ref{sec:dr}). Together, these components enable initialization and iterative optimization of pose and shape in an analysis-by-synthesis framework (Section \ref{sec:inference}).

Our method, SDFEst, takes a cropped point set $\mathcal{P}\subset\mathbb{R}^3$ of the object as input\footnote{To simplify usage of our pipeline, we also integrate our method with Mask R-CNN \cite{he2017mask} from Detectron2 \cite{wu2019detectron2} to enable inference starting from an RGB image $\mathbf{I}$ and depth map $\mathbf{D}$.
}, which we generate from the mask $\mathcal{M}$, depth map $\mathbf{D}$, and the camera projection matrix $\mathbf{P}$. The first part of our method is a novel initialization network, which estimates an initial pose and shape. The shape is predicted as a latent shape descriptor in a low-dimensional shape space learned by a VAE and a separate scalar scaling factor. This initialization gives a coarse estimate which is subsequently refined in an analysis-by-synthesis fashion. To do so, we use the decoder of our VAE to decode the latent shape descriptor into a discretized SDF, render it with a differentiable renderer, compare it with one or more observed depth maps, and compute an optimizable loss to iteratively refine the latent variables.

The embedded generative model is trained with complete shapes, not partial views. Therefore, it decodes the latent shape descriptor, which is inferred from a partial view, to the full reconstructed shape. That is, the pipeline finds the pose, scale, and normalized shape that best matches the observation.

\subsection{Shape Modeling}\label{sec:shape_modeling}
To model per-category shape, we employ a VAE to find a low-dimensional, smooth representation of an object category's shape. The idea is that the VAE constrains the reconstructions to valid shapes from a category and hence automatically completes partial observations when fitting pose and shape to the observations. To represent the shape, we use discretized SDFs which can be easily converted to a mesh using the marching cubes algorithm \cite{lorensen1987marching}.

We follow common practice for training VAEs \cite{kingma2013auto} and jointly train an encoder
\begin{equation}
    \begin{aligned}
        f_\mathrm{enc}: \mathbb{R}^{R\times R\times R} &\to \mathbb{R}^N\times \mathbb{R}_{>0}^N \\
        \mathbf{V}_\mathrm{SDF} &\mapsto (\boldsymbol{\mu},\boldsymbol{\lambda})
    \end{aligned}
\end{equation}
and a decoder
\begin{equation}
    \begin{aligned}
        f_\mathrm{dec}: \quad\mathbb{R}^N &\to \mathbb{R}^{R\times R\times R}\\
        \mathbf{z}_\mathrm{shape} &\mapsto \mathbf{\widetilde{V}}_\mathrm{SDF}.
    \end{aligned}
\end{equation}
Given a discretized SDF $\mathbf{V}_\mathrm{SDF}\in \mathbb{R}^{R\times R\times R}$ of fixed resolution $R$, the encoder predicts the mean $\boldsymbol{\mu}$ and variance $\boldsymbol{\lambda}$ of a multivariate Gaussian $\mathcal{N}(\boldsymbol{\mu}, \mathrm{diag}(\boldsymbol{\lambda}))$ with diagonal covariance in an $N$-dimensional latent space. The decoder reconstructs the SDF given a sample $\mathbf{z}_\mathrm{shape}\sim \mathcal{N}(\boldsymbol{\mu},\mathrm{diag}(\boldsymbol{\lambda}))$ from the latent space. 
During inference in the full pipeline, only the VAE's decoder is used.

As is standard for VAEs, we use Kullback-Leibler divergence to regularize the predictions. When computing the reconstruction error, we aim to give higher priority to correctly capturing the surface instead of the distance in empty space. To do so, we modify a simple reconstruction loss in two ways.

First, following \cite{xu2019}, we increase the weight for distances below a threshold $\delta$ to give higher priority to correctly capturing the surface instead of the distance in empty space. Second, we give additional supervision to the reconstructed surfaces by rendering a depth map of a random view of $\mathbf{V}_\mathrm{SDF}$ and creating a point set $\mathcal{P}$ from it. Applying trilinear interpolation at these points in the reconstructed SDF $\mathbf{\widetilde{V}}_\mathrm{SDF}$ should yield 0. Therefore, we add the sum of these interpolations as an additional loss term.

The total loss function for training the VAE is given by
\begin{equation}
    \begin{aligned}
        L_\mathrm{VAE}=&\lambda_{<\delta} \lVert \mathbf{V}_\mathrm{SDF}^{<\delta} - \mathbf{\widetilde{V}}_\mathrm{SDF}^{<\delta} \rVert_2^2 \\
            & + \lambda_{\geq\delta} \lVert\mathbf{V}_\mathrm{SDF}^{\geq\delta}-\mathbf{\widetilde{V}}_\mathrm{SDF}^{\geq\delta}\rVert_2^2 \\ 
            & + \lambda_\mathrm{SDF} \sum_{\mathbf{p}\in\mathcal{P}} |\mathrm{trilinear}(\mathbf{\widetilde{V}}_\mathrm{SDF}, \mathbf{p})|^2 \\
            & + \lambda_\mathrm{KLD}  D_\mathrm{KL}\left(\mathcal{N}(\boldsymbol{\mu},\boldsymbol{\lambda})\parallel\mathcal{N}(\mathbf{0}, \mathbf{I})\right).
    \end{aligned}
\end{equation}

We tuned the relative importance of these loss terms by visual inspection, such that unconditioned samples $\mathbf{z}_\mathrm{shape}\sim \mathcal{N}(\mathbf{0}, \mathbf{I})$ gave good reconstructions. %

\subsection{Initialization Network}\label{sec:init}
For each object category, we train an initialization network
\begin{equation}
    \begin{aligned}
        f_\mathrm{init}: \mathbb{R}^{M\times 3} &\to \mathbb{R}^3\times \mathbb{R} \times \mathbb [0,1]^G \times \mathbb{R}^N \\
        \mathcal{P} &\mapsto (\mathbf{z}_\mathrm{pos}, z_\mathrm{scale}, \mathbf{o}, \mathbf{z}_\mathrm{shape}).
    \end{aligned}
\end{equation}
Given a point set $\mathcal{P}$ of variable size $M$ (determined by the number of depth pixels within the object mask), the network predicts the position $\mathbf{z}_\mathrm{pos}$, the scale $z_\mathrm{scale}$, a distribution over orientations $\mathbf{o}$, and the latent shape vector $\mathbf{z}_\mathrm{shape}$.

To handle ambiguous sensor data, we use a probabilistic orientation representation by discretizing $\mathrm{SO}(3)$ and predicting the probability of the orientation being inside each grid cell. We use the base grid as specified by \cite{yershova2010generating} since it uniformly discretizes $\mathrm{SO}(3)$ and allows $\mathcal{O}(1)$ conversion from a continuous orientation to the corresponding index. While this introduces a discretization error, it allows the network to represent uncertainty and arbitrary distributions on $\mathrm{SO}(3)$, which is important to handle objects with any (potentially view-dependent) symmetries using the same framework. In Section \ref{section:ablation_study} we further show that, if multiple views are available, this probabilistic output can be used to identify the most certain initial orientation. Specifically, let $g: \mathrm{SO}(3) \to \{1,...,G\}$ denote the mapping from an orientation (we use unit quaternions for $\mathbf{z}_\mathrm{orientation}$) to the index of the $\mathrm{SO}(3)$ grid with $G$ cells. Furthermore, let $h$ denote the inverse operation. Note that in general $h(g(\mathbf{q}))\neq\mathbf{q}$, since a discretization error is introduced and unit quaternions are a double cover of $\mathrm{SO}(3)$. The discretization error that this representation introduces in the initialization of the orientation will subsequently be removed during the optimization which operates on quaternions. %

We use a PointNet-like architecture \cite{qi2017pointnet} and train it on synthetic point sets generated by sampling shapes from the VAE and randomizing their pose and scale. To avoid introducing any bias towards upright objects we use uniform distributions for position and orientation. However, we include a prior on possible sizes of an object category by specifying a distribution for the scale during training. We normalize the masked point set to have zero mean before passing it to the network. That is, the network's position output is only the offset from the masked point set's mean, which we will not denote explicitly in the following. Given such samples, we train the initialization network in a supervised manner with the following loss function:
\begin{equation}
    \begin{aligned}
        L_\mathrm{init}=&\lambda_\mathrm{pos}\lVert\mathbf{z}_\mathrm{pos} - \mathbf{\hat{z}}_\mathrm{pos}\rVert_2^2 \\
            & + \lambda_\mathrm{orientation}(-\log(o_{g(\mathbf{\hat{z}}_\mathrm{orientation})})\\
            & + \lambda_\mathrm{scale}(z_\mathrm{scale} - \hat{z}_\mathrm{scale})^2\\
            & + \lambda_\mathrm{shape}\lVert\mathbf{z}_\mathrm{shape} - \mathbf{\hat{z}}_\mathrm{shape}\rVert_2^2\\
    \end{aligned}
\end{equation}
where $\hat{\cdot}$ indicates the sampled quantities \hln{ and $o_i$ denotes the $i\text{th}$ element of $\mathbf{o}$}. Since real masks will typically not perfectly align with the depth image, and RGB-D sensors exhibit noise close to object edges, we augment the depth images prior to converting to a point set. We found this step to be crucial for usability on real-world data, where robust outlier rejection is difficult, depending on the object's shape and camera angle.

\subsection{Differentiable Renderer}\label{sec:dr}
To enable analysis-by-synthesis optimization with the SDF we render the depth map using a differentiable renderer. We follow the idea of SDFDiff \cite{jiang2020sdfdiff} and apply sphere tracing to quickly find the zero crossing in the SDF and use trilinear interpolation at the last step to compute derivatives with respect to pose, scale, and SDF. %

\subsection{Inference}\label{sec:inference}
The input to our network is the masked point set $\mathcal{P}$ in the camera frame, which is passed to the initialization network to obtain an initial position $\mathbf{z}_\mathrm{pos}$, discretized orientation distribution $\mathbf{o}$, scale $z_\mathrm{scale}$, and latent shape descriptor $\mathbf{z}_\mathrm{shape}$. The discretized orientation distribution $\mathbf{o}$ is converted to a continuous unit quaternion $\mathbf{z}_\mathrm{orientation}=h\left(\argmax_i o_i\right)$. Since we never regress the quaternion directly, we do not suffer from the discontinuity issue discussed in \cite{zhou2019continuity}.

\begin{figure}[htb]
    \centering
    \subfloat[SDF]{
	\input{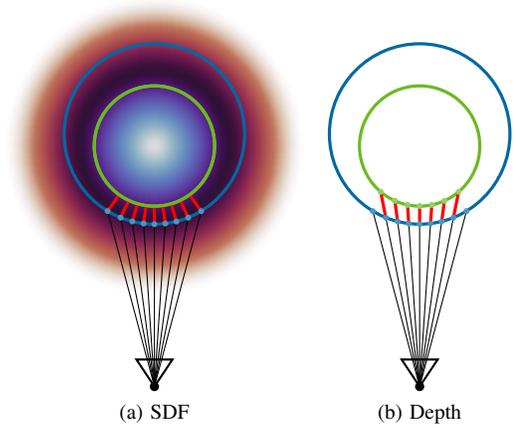}
    }
    \subfloat[Depth]{
	\begin{tikzpicture}[scale=0.8,rotate=90]
    \definecolor{leorange}{rgb}{1,0.27,0}
    \definecolor{leoblue}{rgb}{0,0.41,0.64}
    \definecolor{leogreen}{rgb}{0.42,0.73,0.13}

    \coordinate (camera) at (-4.0, 0);
    \coordinate (virtual) at (3, 1.2);
    \coordinate (object) at (3, 0.75);
    \coordinate (grid) at (2.5, 0.5);

    \draw[name path=estc, very thick, leogreen] (0,0) circle (1.0);

    \draw[name path=obsc, very thick, leoblue] (0.2,0) circle (1.5);

    \path[draw, thick] (camera) --++ (0.45,0.3) --++ (0, -0.6) -- cycle;
    \draw[very thick, fill] (camera) circle (0.05);

    \path[name path=r1] (camera) --++ (6,0);
    \path[name path=r2] (camera) --++ (6,0.4);
    \path[name path=r3] (camera) --++ (6,0.8);
    \path[name path=r4] (camera) --++ (6,1.2);
    \path[name path=r5] (camera) --++ (6,1.6);
    \path[name path=r6] (camera) --++ (6,-0.4);
    \path[name path=r7] (camera) --++ (6,-0.8);
    \path[name path=r8] (camera) --++ (6,-1.2);
    \path[name path=r9] (camera) --++ (6,-1.6);

    \draw[name intersections={of=r1 and obsc}] (camera) -- (intersection-2) coordinate (obs1);
    \draw[name intersections={of=r2 and obsc}] (camera) -- (intersection-2) coordinate (obs2);
    \draw[name intersections={of=r3 and obsc}] (camera) -- (intersection-2) coordinate (obs3);
    \draw[name intersections={of=r4 and obsc}] (camera) -- (intersection-2) coordinate (obs4);
    \draw[name intersections={of=r5 and obsc}] (camera) -- (intersection-2) coordinate (obs5);
    \draw[name intersections={of=r6 and obsc}] (camera) -- (intersection-1) coordinate (obs6);
    \draw[name intersections={of=r7 and obsc}] (camera) -- (intersection-1) coordinate (obs7);
    \draw[name intersections={of=r8 and obsc}] (camera) -- (intersection-1) coordinate (obs8);
    \draw[name intersections={of=r9 and obsc}] (camera) -- (intersection-1) coordinate (obs9);

    \path[name path=e1] (obs1) -- (0,0);
    \path[name path=e2] (obs2) -- (0,0);
    \path[name path=e3] (obs3) -- (0,0);
    \path[name path=e4] (obs4) -- (0,0);
    \path[name path=e5] (obs5) -- (0,0);
    \path[name path=e6] (obs6) -- (0,0);
    \path[name path=e7] (obs7) -- (0,0);
    \path[name path=e8] (obs8) -- (0,0);
    \path[name path=e9] (obs9) -- (0,0);

    \draw[name intersections={of=r1 and estc},red,very thick] (obs1) -- (intersection-2) coordinate (est1);
    \draw[name intersections={of=r2 and estc},red,very thick] (obs2) -- (intersection-2) coordinate (est2);
    \draw[name intersections={of=r3 and estc},red,very thick] (obs3) -- (intersection-2) coordinate (est3);
    \draw[name intersections={of=r4 and estc},red,very thick] (obs4) -- (intersection-2) coordinate (est4);
    \draw[name intersections={of=r6 and estc},red,very thick] (obs6) -- (intersection-1) coordinate (est6);
    \draw[name intersections={of=r7 and estc},red,very thick] (obs7) -- (intersection-1) coordinate (est7);
    \draw[name intersections={of=r8 and estc},red,very thick] (obs8) -- (intersection-1) coordinate (est8);

    \draw[fill, leoblue!70] (obs1) circle (0.04);
    \draw[fill, leoblue!70] (obs2) circle (0.04);
    \draw[fill, leoblue!70] (obs3) circle (0.04);
    \draw[fill, leoblue!70] (obs4) circle (0.04);
    \draw[fill, leoblue!70] (obs5) circle (0.04);
    \draw[fill, leoblue!70] (obs6) circle (0.04);
    \draw[fill, leoblue!70] (obs7) circle (0.04);
    \draw[fill, leoblue!70] (obs8) circle (0.04);
    \draw[fill, leoblue!70] (obs9) circle (0.04);

    \draw[fill, leogreen!70] (est1) circle (0.04);
    \draw[fill, leogreen!70] (est2) circle (0.04);
    \draw[fill, leogreen!70] (est3) circle (0.04);
    \draw[fill, leogreen!70] (est4) circle (0.04);
    \draw[fill, leogreen!70] (est6) circle (0.04);
    \draw[fill, leogreen!70] (est7) circle (0.04);
    \draw[fill, leogreen!70] (est8) circle (0.04);

\end{tikzpicture}
    }
    \caption{Losses used to optimize the pose, scale, and shape of the object. The blue circle is the real observed object, green visualizes the current estimate, and red visualizes the losses. Note that each loss term captures different cases of misalignments, while neither by itself handles all.}\label{fig:losses}
\end{figure}

Starting from this initial latent estimate, we decode the SDF, render it at the current pose, and compute a loss based on the observations. We combine two complementary losses, which are visualized in Fig.~\ref{fig:losses}.

The SDF loss is computed by transforming the observed point set $\mathcal{P}$ into the object frame and interpolating the discretized SDF at the observed points. Since the observed points should lie on the object's surface, the interpolation should be close to $0$. Therefore, we use the following loss:
\begin{equation}\label{eq_lsdf}
    L_\mathrm{SDF}=\frac{1}{|\mathcal{P}|}\sum_{{}^\mathrm{c}\mathbf{p}\in\mathcal{P}}|\mathrm{trilinear}(\mathbf{\widetilde{V}}_\mathrm{SDF},z_\mathrm{scale}^{-1}\tensor[^{\mathrm{o}}]{\widetilde{\mathbf{T}}}{_{\mathrm{w}}}\tensor[^{\mathrm{w}}]{\mathbf{T}}{_{\mathrm{c}}}\tensor[^{\mathrm{c}}]{\mathbf{p}}{})|,
\end{equation}
where $\tensor[^{\mathrm{o}}]{\widetilde{\mathbf{T}}}{_{\mathrm{w}}}$ is the 6D transformation computed from $\mathbf{z}_\mathrm{pose}$.

The depth loss is computed based on the observed depth map $\mathbf{D}$ and estimated depth map $\widetilde{\mathbf{D}}$ as
\begin{equation}
    L_\mathrm{depth}=\frac{1}{|\widetilde{\mathcal{D}}|}\sum_{(i,j)\in\widetilde{\mathcal{D}}}| \widetilde{D}_{i,j} - D_{i,j}|,
\end{equation}
where $\widetilde{\mathcal{D}}=\{(i,j)\subseteq\mathcal{M}\mid D_{i,j}\neq 0 \land \widetilde{D}_{i,j}\neq 0\}$, that is, the set of pixels where both the masked depth map and the rendered depth map are valid.

The total loss is then computed as 
\begin{equation}\label{eq_l}
    L = \lambda_\mathrm{SDF} L_\mathrm{SDF} + \lambda_\mathrm{depth} L_\mathrm{depth}.
\end{equation}
Since the whole pipeline is differentiable, we can use any first-order optimization algorithm to jointly optimize pose, scale, and shape of the object.

\hln{Our framework readily allows incorporating additional information from multiple views by summing up the per-view losses. Specifically, when $K$ views and their poses $\tensor[^{\mathrm{w}}]{\mathbf{T}}{_{\mathrm{c}k}}, k=1,...,K$ are available, we evaluate \eqref{eq_lsdf}-\eqref{eq_l} to retrieve per-view losses $L_k$ and compute the total loss as $L=\sum_{k=1}^K L_k$.}

\subsection{Training Details}\label{sec:training_details}
We train a separate VAE and initialization network for each object category. Our implementation is based on PyTorch \cite{paszke2019pytorch} and Open3D \cite{zhou2018open3d}. Both networks are trained using Adam optimizer \cite{kingma2014adam} with a learning rate of \num{1e-3}.

\subsubsection{VAE Training}\label{sec:vae_training}
We use CAD models from the ShapeNet dataset \cite{chang2015}, convert them to SDFs with a resolution of $R=64$, and manually remove erroneous CAD models (i.e.,~conversion to SDF failed, CAD model is misclassified / contains multiple objects / etc.) prior to training. The exact subsets are published as part of our open-source software. 

To stabilize the VAE training, we set $\lambda_\mathrm{KLD}=0$ for the first 1000 iterations of training. This helped to avoid local minima where the encoder predicts $\boldsymbol{\mu}\approx \mathbf{0}$, and the decoder predicts only a constant (mean) value.

\subsubsection{Initialization Network Training}
To train the initialization network, we need to generate single-view point sets as close as possible to the expected preprocessed real-world point sets captured by an RGB-D camera. To do so, we sample $\mathbf{z}_\mathrm{shape}\sim\mathcal{N}(\mathbf{0}, \mathbf{I})$, that is, from the prior distribution of the VAE. To generate random object positions, we first uniformly sample a pixel in the image plane (we generate 640x480 depth maps) to compute the ray on which the SDF center will be. We then sample a typical object distance along that ray to define the 3D position of the center. Next, we sample the scale of the SDF from per-category specified distributions. Finally, we uniformly sample a quaternion representing the orientation of the object. To robustify our network to typical outliers observed for noisy masks, we apply a random affine transformation to the mask and generate a uniform outlier value for the parts where the noisy mask does not overlap with the rendered depth image anymore. With a probability of 0.5, we further apply a Gaussian filter to the previously augmented depth image. We found this to be a simple way of simulating flying pixels at object boundaries commonly observed with RGB-D cameras.
We train the initialization network by generating the data on the fly. %

\section{Experiments}

In this section, we will discuss four experiments. First, we follow our proposed evaluation protocol of pose and shape estimation from real single-view data \cite{bruns2022evaluation}. Second, we follow the evaluation protocol used by NodeSLAM \cite{sucar2020nodeslam} to compare pose and shape estimation with multiple views. Third, we perform an ablation study to assess the effect of the different components. Finally, we provide a run time analysis of our method.

Unless otherwise stated, we use the same networks \hln{($N=8, G=576$)} trained on synthetic data only and optimize for 50 iterations, which is typically sufficient for convergence.

\subsection{Single-view Real Data}\label{sec:realdata}

For all real-data experiments, we follow the protocol and evaluation metrics defined in \cite{bruns2022evaluation}. To summarize, we report precision (i.e., \# correct estimates / \# total estimates) based on varying thresholds on position, orientation, and $F_{1\mathrm{cm}}$-score. As baselines, we use CASS \cite{chen2020learning}, SPD \cite{tian2020shape}, ASM-Net \cite{akizuki2021}, and iCaps \cite{deng2022icaps}. For a fair comparison of pose and shape estimation, we use the same ground-truth masks and categories for all methods as described in \cite{bruns2022evaluation}. In Table \ref{tab:real275} and Table \ref{tab:redwood75} we show results on REAL275 \cite{wang2019normalized} and REDWOOD75 \cite{choi2016large,bruns2022evaluation}, respectively. The green highlighted methods have only access to synthetic data. 

On REAL275 (Table \ref{tab:real275}), our method performs better than CASS, ASM-Net, and iCaps for all thresholds and on par with SPD. We further show results where our method is constrained to orientations present in the REAL training data (denoted by Ours$^\dagger$). With this modification, our method outperforms all other methods on REAL275, at the cost of poorer generalization for unconstrained orientations. 

\begin{table}[t!]
    \centering
    \begin{threeparttable}
        \caption{REAL275 results}\label{tab:real275}
        \begin{tabular}{lrrBBBr}
            \hlineB{2}
            \TBstrut & CASS & SPD & ASM & \hln{iCaps} & Ours & Ours$\smash{^\dagger}$ \\
            \hline
            \Tstrut 10\degree,\SI{2}{cm} & 0.331 & 0.535 & 0.331 & \hln{0.205} & 0.506 & \textbf{0.589} \\
            5\degree,\SI{1}{cm} & 0.073 & 0.205 & 0.069 & \hln{0.030} & 0.224 & \textbf{0.242}\\
            10\degree,\SI{2}{cm},0.6 & 0.031 & 0.471 & 0.215 & \hln{0.106} & 0.442 & \textbf{0.491}\\
            \Bstrut 5\degree,\SI{1}{cm},0.8 & 0.000 & 0.170 & 0.050 &  \hln{0.013} & 0.185 & \textbf{0.191} \\
            \hlineB{2}
        \end{tabular}
        \begin{tablenotes}
            \item[$^\dagger$] Initialization constrained to orientations in REAL train split
        \end{tablenotes}
    \end{threeparttable}
    
    \raggedright
    \scriptsize
\end{table}

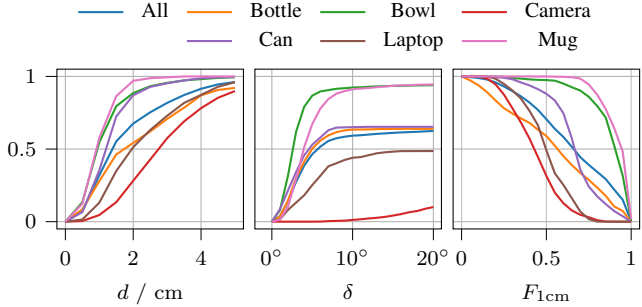
\begin{figure}[t!]
    \centering%
    \begin{tikzpicture}

\definecolor{crimson2143940}{RGB}{214,39,40}
\definecolor{darkgray176}{RGB}{176,176,176}
\definecolor{darkorange25512714}{RGB}{255,127,14}
\definecolor{forestgreen4416044}{RGB}{44,160,44}
\definecolor{gray127}{RGB}{127,127,127}
\definecolor{lightgray204}{RGB}{204,204,204}
\definecolor{mediumpurple148103189}{RGB}{148,103,189}
\definecolor{orchid227119194}{RGB}{227,119,194}
\definecolor{sienna1408675}{RGB}{140,86,75}
\definecolor{steelblue31119180}{RGB}{31,119,180}

\begin{axis}[%
legend columns=4,
scale only axis,width=1mm, %
hide axis,
legend image post style={sharp plot},
legend style={draw=none,column sep=1ex,font=\footnotesize}
]

\addplot [thick, steelblue31119180]
table {%
0 0
1 1
};
\addlegendentry{All}

\addplot [thick, darkorange25512714]
table {%
0 0
1 1
};
\addlegendentry{Bottle}

\addplot [thick, forestgreen4416044]
table {%
0 0
1 1
};
\addlegendentry{Bowl}

\addplot [thick, crimson2143940]
table {%
0 0
1 1
};
\addlegendentry{Camera}

\addlegendimage{empty legend}
\addlegendentry{}

\addplot [thick, mediumpurple148103189]
table {%
0 0
1 1
};
\addlegendentry{Can}

\addplot [thick, sienna1408675]
table {%
0 0
1 1
};
\addlegendentry{Laptop}

\addplot [thick, orchid227119194]
table {%
0 0
1 1
};
\addlegendentry{Mug}
\end{axis}%
\end{tikzpicture}%
%
    
    \hspace{-0.08\linewidth}%
    \begin{tikzpicture}%

\definecolor{crimson2143940}{RGB}{214,39,40}
\definecolor{darkgray176}{RGB}{176,176,176}
\definecolor{darkorange25512714}{RGB}{255,127,14}
\definecolor{forestgreen4416044}{RGB}{44,160,44}
\definecolor{gray127}{RGB}{127,127,127}
\definecolor{lightgray204}{RGB}{204,204,204}
\definecolor{mediumpurple148103189}{RGB}{148,103,189}
\definecolor{orchid227119194}{RGB}{227,119,194}
\definecolor{sienna1408675}{RGB}{140,86,75}
\definecolor{steelblue31119180}{RGB}{31,119,180}

\begin{axis}[
width=0.28\linewidth,
height=0.24\linewidth,
scale only axis=true,
tick align=outside,
tick pos=left,
x grid style={darkgray176},
xlabel={$d\vphantom{ / \mathrm{cm}}$},
xmajorgrids,
xmin=-0.0025, xmax=0.0525,
xtick style={color=black},
xticklabel=$\mathclap{\pgfmathprintnumber{\tick}}\vphantom{\degree{}}$,
y grid style={darkgray176},
ymajorgrids,
ymin=-0.05, ymax=1.05,
ytick style={color=black},
change x base,
unit markings=slash space,
x SI prefix=centi,x unit=m,
every x tick scale label/.append style={overlay}
]
\addplot [thick, steelblue31119180]
table {%
0 0
0.005 0.0702940811515076
0.01 0.319456508251644
0.015 0.552736071472888
0.02 0.675393969475121
0.025 0.752884973321752
0.03 0.817533192703809
0.035 0.871944409976424
0.04 0.914877776399057
0.045 0.94509244323117
0.05 0.961471646606279
};
\addplot [thick, darkorange25512714]
table {%
0 0
0.005 0.0845121508886471
0.01 0.279289082335872
0.015 0.462459194776931
0.02 0.542256075444324
0.025 0.620602103735945
0.03 0.707290533188248
0.035 0.784911135291984
0.04 0.869423286180631
0.045 0.906782734856728
0.05 0.920928545520493
};
\addplot [thick, forestgreen4416044]
table {%
0 0
0.005 0.133755274261603
0.01 0.546835443037975
0.015 0.79493670886076
0.02 0.885232067510549
0.025 0.932489451476793
0.03 0.955696202531646
0.035 0.972573839662447
0.04 0.981434599156118
0.045 0.989873417721519
0.05 0.994092827004219
};
\addplot [thick, crimson2143940]
table {%
0 0
0.005 0.00449606594230049
0.01 0.0457100037467216
0.015 0.134132633945298
0.02 0.281753465717497
0.025 0.427126264518546
0.03 0.574372424128887
0.035 0.690520794304983
0.04 0.78119145747471
0.045 0.853128512551517
0.05 0.899213188460097
};
\addplot [thick, mediumpurple148103189]
table {%
0 0
0.005 0.0663246610480029
0.01 0.353242946134115
0.015 0.724441187248076
0.02 0.870282154635398
0.025 0.92781238548919
0.03 0.952729937706119
0.035 0.972150971051667
0.04 0.987541223891535
0.045 0.994503481128619
0.05 0.996335654085746
};
\addplot [thick, sienna1408675]
table {%
0 0
0.005 0.0148874364560639
0.01 0.130718954248366
0.015 0.348946986201888
0.02 0.513071895424837
0.025 0.630718954248366
0.03 0.729121278140886
0.035 0.817356572258533
0.04 0.872912127814089
0.045 0.928467683369644
0.05 0.960058097312999
};
\addplot [thick, orchid227119194]
table {%
0 0
0.005 0.122930609369496
0.01 0.57661148291652
0.015 0.864388869320183
0.02 0.969707643536457
0.025 0.988023952095808
0.03 0.99365973934484
0.035 0.999647763296935
0.04 1
0.045 1
0.05 1
};
\end{axis}

\end{tikzpicture}%
%
    \hspace{-0.1\linewidth}%
    \hfil%
    \begin{tikzpicture}%

\definecolor{crimson2143940}{RGB}{214,39,40}
\definecolor{darkgray176}{RGB}{176,176,176}
\definecolor{darkorange25512714}{RGB}{255,127,14}
\definecolor{forestgreen4416044}{RGB}{44,160,44}
\definecolor{gray127}{RGB}{127,127,127}
\definecolor{lightgray204}{RGB}{204,204,204}
\definecolor{mediumpurple148103189}{RGB}{148,103,189}
\definecolor{orchid227119194}{RGB}{227,119,194}
\definecolor{sienna1408675}{RGB}{140,86,75}
\definecolor{steelblue31119180}{RGB}{31,119,180}

\begin{axis}[
width=0.28\linewidth,
height=0.24\linewidth,
scale only axis=true,
tick align=outside,
tick pos=left,
x grid style={darkgray176},
xlabel={$\delta\vphantom{ / \mathrm{cm}}$},
xticklabel=$\mathclap{\pgfmathprintnumber{\tick}\degree{}}$,
yticklabels={,,}.
xmin=-1, xmax=21,
xtick style={color=black},
y grid style={darkgray176},
grid=both,
ymin=-0.05, ymax=1.05,
ytick style={draw=none}
]
\addplot [thick, steelblue31119180]
table {%
0 0
1 0.0374115895272366
2 0.1400918228068
3 0.274351656533069
4 0.385469661248294
5 0.46426355627249
6 0.516503288249162
7 0.555962278198288
8 0.574450924432312
9 0.584687926541755
10 0.592195061422013
11 0.595173098399305
12 0.599826281176325
13 0.604541506390371
14 0.607829755552798
15 0.61142821690036
16 0.614282169003598
17 0.616763866484676
18 0.619059436654672
19 0.621851346320883
20 0.624208958927907
};
\addplot [thick, darkorange25512714]
table {%
0 0
1 0.0359085963003264
2 0.145085237577077
3 0.300689154878491
4 0.413130214000725
5 0.501632208922742
6 0.557852738483859
7 0.594124047878128
8 0.616249546608633
9 0.628581791802684
10 0.634022488211824
11 0.63474791439971
12 0.635473340587595
13 0.637286906057309
14 0.638012332245194
15 0.638737758433079
16 0.638737758433079
17 0.639100471527022
18 0.639100471527022
19 0.639100471527022
20 0.639463184620965
};
\addplot [thick, forestgreen4416044]
table {%
0 0
1 0.0789029535864979
2 0.317721518987342
3 0.618565400843882
4 0.790717299578059
5 0.867510548523207
6 0.89662447257384
7 0.905907172995781
8 0.914767932489451
9 0.918987341772152
10 0.923628691983122
11 0.925316455696203
12 0.928691983122363
13 0.931645569620253
14 0.932911392405063
15 0.935021097046413
16 0.935443037974684
17 0.936708860759494
18 0.937974683544304
19 0.938818565400844
20 0.939662447257384
};
\addplot [thick, crimson2143940]
table {%
0 0
1 0
2 0
3 0
4 0
5 0
6 0.000749344323716748
7 0.00374672161858374
8 0.00524541026601723
9 0.0071187710753091
10 0.011989509179468
11 0.0164855751217685
12 0.0221056575496441
13 0.0266017234919445
14 0.0329711502435369
15 0.0419632821281379
16 0.052079430498314
17 0.0648182840014987
18 0.0734357437242413
19 0.0887973023604346
20 0.100786811539903
};
\addplot [thick, mediumpurple148103189]
table {%
0 0
1 0.0857456943935507
2 0.215097105166728
3 0.341517039208501
4 0.456211066324661
5 0.529497984609747
6 0.579333089043606
7 0.62953462806889
8 0.647489923048736
9 0.649322096005863
10 0.650787834371565
11 0.651887138145841
12 0.652986441920117
13 0.652986441920117
14 0.652986441920117
15 0.652986441920117
16 0.652986441920117
17 0.652986441920117
18 0.652986441920117
19 0.652986441920117
20 0.652986441920117
};
\addplot [thick, sienna1408675]
table {%
0 0
1 0.0159767610748003
2 0.0809731299927378
3 0.131445170660857
4 0.181554103122731
5 0.251633986928105
6 0.315541031227306
7 0.38162672476398
8 0.402323892519971
9 0.42483660130719
10 0.440087145969499
11 0.44480755265069
12 0.457879448075526
13 0.470951343500363
14 0.477487291212781
15 0.482207697893972
16 0.485838779956427
17 0.485838779956427
18 0.486201888162672
19 0.486201888162672
20 0.486201888162672
};
\addplot [thick, orchid227119194]
table {%
0 0
1 0.0137372314195139
2 0.103909827404015
3 0.293413173652695
4 0.512504402958788
5 0.671363156040859
6 0.778443113772455
7 0.844311377245509
8 0.881648467770342
9 0.8989080662205
10 0.911588587530821
11 0.916519901373723
12 0.920394505107432
13 0.926030292356464
14 0.930609369496301
15 0.935540683339204
16 0.93835857696372
17 0.939063050369848
18 0.942585417400493
19 0.943289890806622
20 0.944346600915815
};
\end{axis}

\end{tikzpicture}%
%
    \hspace{-0.1\linewidth}%
    \hfil%
    \begin{tikzpicture}%

\definecolor{crimson2143940}{RGB}{214,39,40}
\definecolor{darkgray176}{RGB}{176,176,176}
\definecolor{darkorange25512714}{RGB}{255,127,14}
\definecolor{forestgreen4416044}{RGB}{44,160,44}
\definecolor{gray127}{RGB}{127,127,127}
\definecolor{lightgray204}{RGB}{204,204,204}
\definecolor{mediumpurple148103189}{RGB}{148,103,189}
\definecolor{orchid227119194}{RGB}{227,119,194}
\definecolor{sienna1408675}{RGB}{140,86,75}
\definecolor{steelblue31119180}{RGB}{31,119,180}

\begin{axis}[
width=0.28\linewidth,
height=0.24\linewidth,
scale only axis=true,
tick align=outside,
tick pos=left,
yticklabels={,,}
x grid style={darkgray176},
xticklabel=$\mathclap{\pgfmathprintnumber{\tick}}\vphantom{\degree{}}$,
xlabel={$F_{1\mathrm{cm}}\vphantom{ / \mathrm{cm}}$},
xmajorgrids,
xmin=-0.05, xmax=1.05,
xtick style={color=black},
y grid style={darkgray176},
ymajorgrids,
ymin=-0.05, ymax=1.05,
ytick style={color=black},
ytick style={draw=none}
]
\addplot [thick, steelblue31119180]
table {%
0 1
0.05 0.995222732348927
0.1 0.989018488646234
0.15 0.978533316788683
0.2 0.959114033999256
0.25 0.93076064027795
0.3 0.898374488149895
0.35 0.862700086859412
0.4 0.820883484303263
0.4 0.820883484303263
0.45 0.769822558630103
0.5 0.70548455143318
0.55 0.639161186251396
0.6 0.583695247549324
0.65 0.515572651693759
0.7 0.444223849112793
0.75 0.391115523017744
0.8 0.342412209951607
0.85 0.292964387641147
0.9 0.220809033378831
0.95 0.150701079538404
1 0.00397071596972329
};
\addplot [thick, darkorange25512714]
table {%
0 1
0.05 0.973159231048241
0.1 0.938701487123685
0.15 0.891911498005078
0.2 0.834602829162133
0.25 0.781284004352557
0.3 0.744287268770403
0.35 0.710917664127675
0.4 0.680449764236489
0.4 0.680449764236489
0.45 0.633297062023939
0.5 0.589771490750816
0.55 0.523032281465361
0.6 0.451577801958651
0.65 0.389916575988393
0.7 0.335872324990932
0.75 0.287631483496554
0.8 0.234312658686979
0.85 0.177729416031919
0.9 0.104824084149438
0.95 0.073993471164309
1 0
};
\addplot [thick, forestgreen4416044]
table {%
0 1
0.05 1
0.1 1
0.15 1
0.2 0.99915611814346
0.25 0.99620253164557
0.3 0.988607594936709
0.35 0.984810126582279
0.4 0.979746835443038
0.4 0.979746835443038
0.45 0.976371308016878
0.5 0.974683544303797
0.55 0.970464135021097
0.6 0.951054852320675
0.65 0.925738396624473
0.7 0.90253164556962
0.75 0.859915611814346
0.8 0.7957805907173
0.85 0.717299578059072
0.9 0.523628691983122
0.95 0.316877637130802
1 0.00717299578059072
};
\addplot [thick, crimson2143940]
table {%
0 1
0.05 0.999250655676283
0.1 0.99775196702885
0.15 0.985387785687523
0.2 0.945672536530536
0.25 0.874484825777445
0.3 0.780442113150993
0.35 0.682652678905957
0.4 0.573248407643312
0.4 0.573248407643312
0.45 0.445859872611465
0.5 0.315099288122892
0.55 0.199700262270513
0.6 0.127388535031847
0.65 0.077182465342825
0.7 0.0475833645560135
0.75 0.0303484451105283
0.8 0.0104908205320345
0.85 0.000749344323716748
0.9 0
0.95 0
1 0
};
\addplot [thick, mediumpurple148103189]
table {%
0 1
0.05 0.999633565408575
0.1 0.999267130817149
0.15 0.998167827042873
0.2 0.994869915720044
0.25 0.990472700622939
0.3 0.980212532063027
0.35 0.958592891168926
0.4 0.938438988640528
0.4 0.938438988640528
0.45 0.912422132649322
0.5 0.88054232319531
0.55 0.834371564675705
0.6 0.747526566507878
0.65 0.573470135580799
0.7 0.358373030414071
0.75 0.24221326493221
0.8 0.17405643092708
0.85 0.120923415170392
0.9 0.0729204836936607
0.95 0.0373763283253939
1 0
};
\addplot [thick, sienna1408675]
table {%
0 1
0.05 1
0.1 1
0.15 0.998547567175018
0.2 0.984749455337691
0.25 0.948075526506899
0.3 0.903413217138707
0.35 0.847494553376906
0.4 0.763616557734205
0.4 0.763616557734205
0.45 0.664488017429194
0.5 0.491285403050109
0.55 0.332607116920842
0.6 0.253449527959332
0.65 0.161946259985476
0.7 0.070079883805374
0.75 0.0177923021060276
0.8 0.000726216412490922
0.85 0
0.9 0
0.95 0
1 0
};
\addplot [thick, orchid227119194]
table {%
0 1
0.05 1
0.1 1
0.15 1
0.2 1
0.25 1
0.3 1
0.35 1
0.4 1
0.4 1
0.45 0.999647763296935
0.5 0.999647763296935
0.55 0.998238816484678
0.6 0.997182106375484
0.65 0.994716449454033
0.7 0.985206058471293
0.75 0.94469883761888
0.8 0.874251497005988
0.85 0.774920746741811
0.9 0.644593166607961
0.95 0.483268756604438
1 0.0165551250440296
};
\end{axis}

\end{tikzpicture}%
%
    \vspace{-0.35cm}
    \caption{Per-category estimation precision on REAL275 for varying position, orientation and $F_{1\mathrm{cm}}$-score thresholds.}
    \label{fig:varying_thresholds}
\end{figure}

To better understand these results, we show the results of our unconstrained method for different categories and varying thresholds in Fig.~\ref{fig:varying_thresholds}. We observe that for cans and laptops the orientation precision of our method saturates at 40-60\%. We find that in both cases, pose is ambiguous from geometry alone, and since our method was trained with a uniform orientation distribution, we cannot always infer the correct orientation. If we constrain\footnote{We use our orientation discretization to change the prior distribution without retraining any network.} the initial orientations to those in the REAL train split, these ambiguities disappear, which explains the improved performance in Table \ref{tab:real275}.

\begin{figure}[tbh]
    \centering
    \footnotesize
    \setlength{\tabcolsep}{0pt}
    \begin{tabular}{C{0.19\linewidth}C{0.19\linewidth}C{0.19\linewidth}C{0.19\linewidth}C{0.19\linewidth}}
        GT & CASS & SPD & ASM-Net & Ours
    \end{tabular}
    
    \includegraphics[trim={4cm 0 4cm 0},clip,width=0.19\linewidth,height=0.19\linewidth]{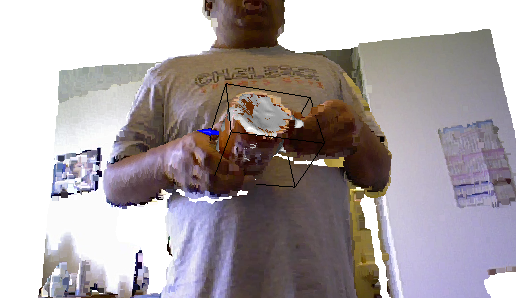}%
    \includegraphics[trim={4cm 0 4cm 0},clip,width=0.19\linewidth,height=0.19\linewidth]{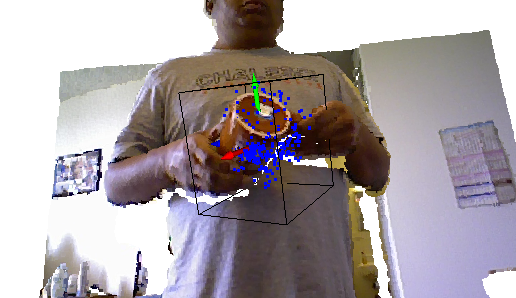}%
    \includegraphics[trim={4cm 0 4cm 0},clip,width=0.19\linewidth,height=0.19\linewidth]{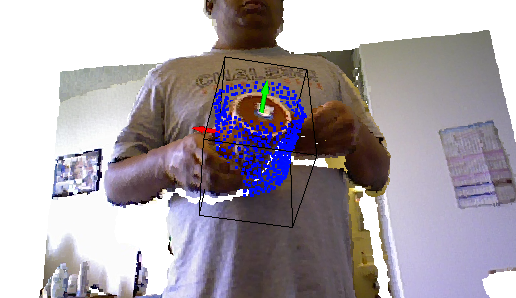}%
    \includegraphics[trim={4cm 0 4cm 0},clip,width=0.19\linewidth,height=0.19\linewidth]{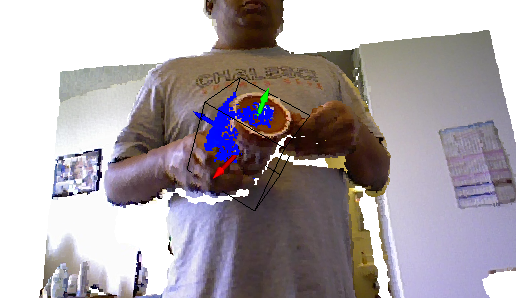}%
    \includegraphics[trim={4cm 0 4cm 0},clip,width=0.19\linewidth,height=0.19\linewidth]{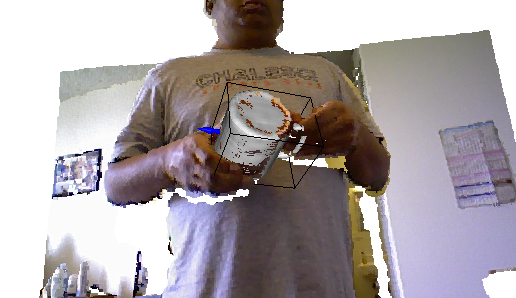}%

    \includegraphics[trim={4cm 0 4cm 0},clip,width=0.19\linewidth,height=0.19\linewidth]{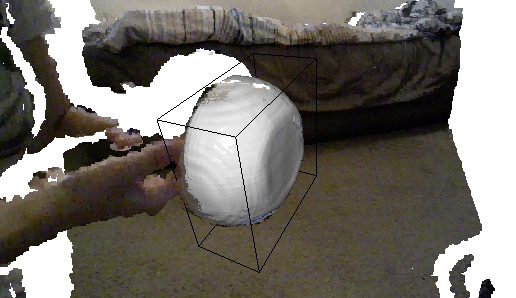}%
    \includegraphics[trim={4cm 0 4cm 0},clip,width=0.19\linewidth,height=0.19\linewidth]{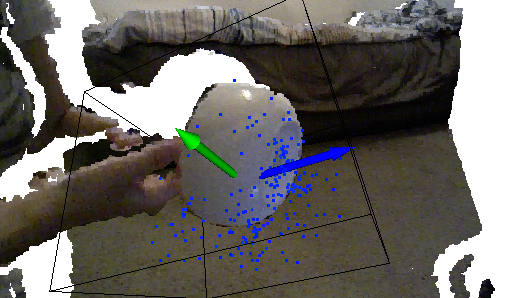}%
    \includegraphics[trim={4cm 0 4cm 0},clip,width=0.19\linewidth,height=0.19\linewidth]{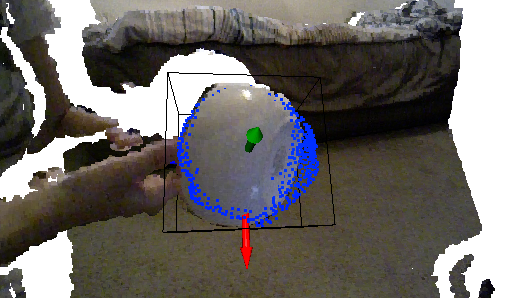}%
    \includegraphics[trim={4cm 0 4cm 0},clip,width=0.19\linewidth,height=0.19\linewidth]{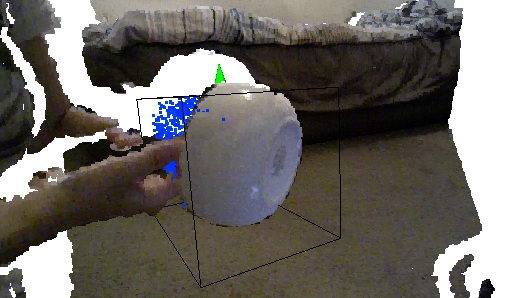}%
    \includegraphics[trim={4cm 0 4cm 0},clip,width=0.19\linewidth,height=0.19\linewidth]{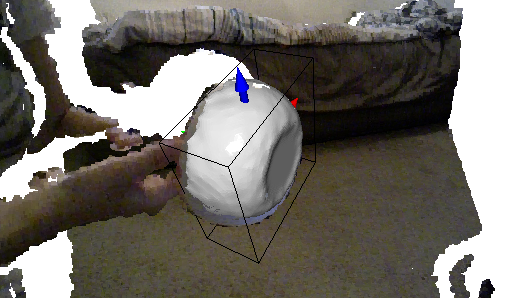}%

    \includegraphics[trim={4cm 0 4cm 0},clip,width=0.19\linewidth,height=0.19\linewidth]{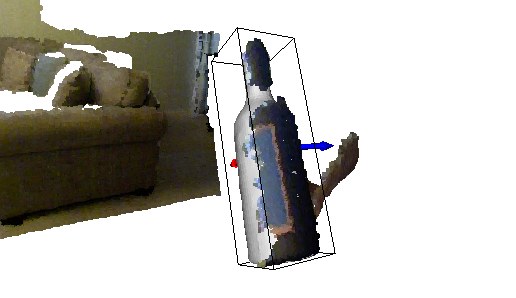}%
    \includegraphics[trim={4cm 0 4cm 0},clip,width=0.19\linewidth,height=0.19\linewidth]{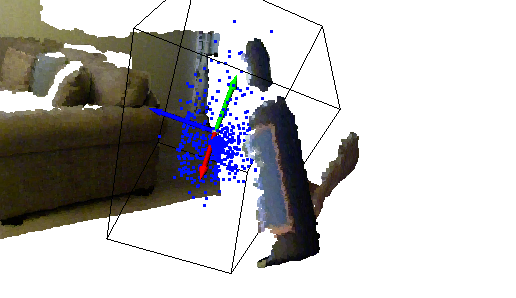}%
    \includegraphics[trim={4cm 0 4cm 0},clip,width=0.19\linewidth,height=0.19\linewidth]{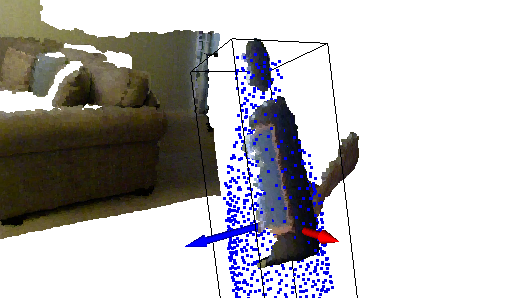}%
    \includegraphics[trim={4cm 0 4cm 0},clip,width=0.19\linewidth,height=0.19\linewidth]{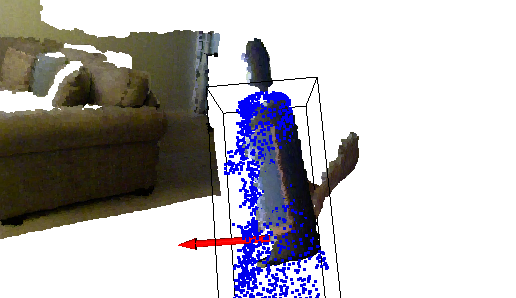}%
    \includegraphics[trim={4cm 0 4cm 0},clip,width=0.19\linewidth,height=0.19\linewidth]{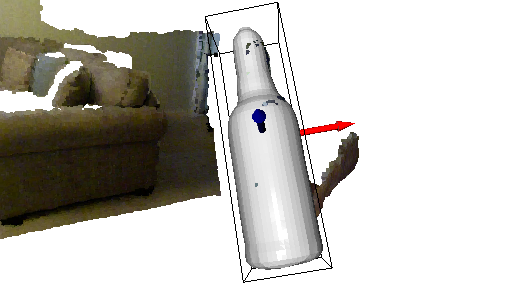}%
    
    \caption{Comparison of the estimated pose and shape on REDWOOD75 data. CASS and SPD, which are trained on datasets that only contain upright objects, fail at estimating unconstrained orientations. The exact training distribution of ASM-Net is unknown. Our method is trained with uniformly distributed orientations and therefore performs significantly better on unconstrained orientations. See Table \ref{tab:redwood75} for quantitative results.}
    \label{fig:redwood75_qualitative}
\end{figure}

On REDWOOD75 (Table \ref{tab:redwood75}), our unconstrained method outperforms the baselines by a large margin. 
To analyze this, 
we show a qualitative comparison for typical REDWOOD75 samples in Fig.~\ref{fig:redwood75_qualitative}. We find that other methods, which were trained on constrained orientations, fail at estimating objects with unconstrained orientations. Our method, on the other hand, only uses synthetic data and, therefore, we were able to generate unconstrained orientations during training.

\begin{table}[tbh]
    \centering
    \begin{threeparttable}
        \caption{REDWOOD75 results}\label{tab:redwood75}
        \begin{tabular}{lrrBBB}
            \hlineB{2}
            \TBstrut  & CASS & SPD & ASM & \hln{iCaps} & Ours \\
            \hline
            \Tstrut 10\degree,\SI{2}{cm} & 0.013 & 0.200 & 0.307 & \hln{0.270} & \textbf{0.653} \\
            5\degree,\SI{1}{cm} & 0.000 & 0.013 & 0.080 & \hln{0.080} & \textbf{0.466} \\
            10\degree,\SI{2}{cm},0.6 & 0.000 & 0.173 & 0.173 & \hln{0.200} & \textbf{0.586} \\
            \Bstrut 5\degree,\SI{1}{cm},0.8 & 0.000 & 0.013 & 0.053 & \hln{0.040} & \textbf{0.413} \\
            \hlineB{2}
        \end{tabular}
    \end{threeparttable}
\end{table}

\subsection{Multi-view Synthetic Data}\label{sec:synthetic}

To quantitatively compare the shape completion and pose estimation capabilities of our approach with NodeSLAM's \cite{sucar2020nodeslam}, without their code available, we attempt to replicate their evaluation procedure: meshes from the ShapeNet \cite{chang2015} mug category are scaled by \num{0.1} (making them approximately \SI{6}{\cm} large), 1, 2, and 3 camera orientations are uniformly sampled and the camera positions are chosen such that the origin of the mesh lies \SI{30}{\cm} in front of each camera on its principal axis. We use Open3D \cite{zhou2018open3d} to render the input images. Note that some details, such as the image resolution and camera parameters, were unknown to us, and the experimental setup might not be completely the same. For this experiment, we only optimize for 30 iterations similar to NodeSLAM. Additionally, we perform 6D pose estimation, whereas \cite{sucar2020nodeslam} only describes orientation estimation around the object's up-axis. We report the same metrics as \cite{sucar2020nodeslam}: reconstruction precision $\mathrm{P}$, chamfer distance $\mathrm{CD}$, thresholded reconstruction precision $\mathrm{P}_{1\mathrm{cm}}$, and thresholded reconstruction recall $\mathrm{R}_{1\mathrm{cm}}$.

The comparison between our results and those reported in \cite{sucar2020nodeslam} is shown in Table \ref{tab:quantitative_comparison}. Our method achieves better results on most metrics. It can be seen that both methods successfully improve the estimates with the additional information provided by additional views. Note that such a multi-view optimization is not readily supported by the single-view methods we compared with in Section \ref{sec:realdata}.

\begin{table}[tbh]
    \centering
    \begin{threeparttable}
        \caption{Multi-view comparison with NodeSLAM \cite{sucar2020nodeslam}}\label{tab:quantitative_comparison}
        \setlength{\tabcolsep}{5pt}
        \begin{tabular}{@{}lrrcrrcrr@{}}
            \toprule
            & \multicolumn{2}{c}{1 view} & & \multicolumn{2}{c}{2 views} & & \multicolumn{2}{c}{3 views} \\
            \cmidrule{2-3} \cmidrule{5-6} \cmidrule{8-9}
            & Ours & \cite{sucar2020nodeslam} & & Ours & \cite{sucar2020nodeslam} & & Ours & \cite{sucar2020nodeslam}\\
            \midrule
            $\mathrm{P}/\mathrm{mm}$ & $ 4.625 $ & $ \mathbf{4.459} $ & & $\mathbf{ 3.290 }$ & $ 3.752 $ & & $\mathbf{ 2.572}$ & $ 3.484 $ \\
            $\mathrm{CD}/\mathrm{mm}$ & $\mathbf{ 3.782 }$ & $ 4.439 $ & & $\mathbf{ 2.818 }$ & $ 3.854 $ & & $\mathbf{ 2.330 }$ & $ 3.648 $ \\
            $\mathrm{P}_{1\mathrm{cm}}/\%$ & $\mathbf{ 86.53 }$ & $ - $ & & $\mathbf{ 92.80 }$ & $ - $ & & $\mathbf{ 96.40 }$ & $ - $ \\
            $\mathrm{R}_{1\mathrm{cm}}/\%$ & $\mathbf{ 94.55 }$ & $ 93.49 $ & & $\mathbf{ 96.62 }$ & $ 95.72 $ & & $\mathbf{ 97.53 }$ & $ 96.07 $ \\
            \bottomrule
        \end{tabular}
    \end{threeparttable}
\end{table}

\subsection{Ablation Study}\label{section:ablation_study}

To verify the effectiveness of different components in our pipeline, we performed an ablation study using the 3-view experimental setup described in Section \ref{sec:synthetic}. Table \ref{tab:ablation} summarizes the results. First, instead of initializing with a random first view, we assume that all views are available and initialize based on the view that gives the highest probability in the orientation distribution (\emph{best view}). It can be seen that this simple change further improves the results. Qualitatively, we found that unambiguous views score reliably higher than ambiguous ones, which is consistent with the improved results. \emph{Depth loss only} and \emph{SDF loss only} show that none of the losses are sufficient by themselves, which supports the discussion in Section \ref{sec:inference}. We further investigate the importance of initialization and iterative optimization. \emph{Init only} shows that skipping the optimization completely gives significantly worse results, which is expected due to the quantization error that is introduced due to the $\mathrm{SO}(3)$ discretization. Skipping only the shape optimization (\emph{no shape opt.}) also gives worse results, but to a lesser degree. Initializing the shape to the mean shape $\mathbf{z}_\mathrm{shape}=\mathbf{0}$, but optimizing it, similar to \cite{sucar2020nodeslam}, also gives significantly worse results. \hln{Finally, we compare our discretized $\mathrm{SO}(3)$ grid to regressing a single \emph{quaternion} and using a \emph{finer grid}. Both perform worse. For quaternions, the most likely reason for the drop in performance is the inability of handling multimodal distributions. The finer grid resolution converged slower during training than the coarser one we used for our experiments. We believe that training for a longer time might close the performance gap. %
}

\begin{table}[tbh]
    \centering
    \begin{threeparttable}
        \caption{Ablation study}\label{tab:ablation}
        \begin{tabular}{@{}lrrrr@{}}
            \toprule
            & $\mathrm{P}/\mathrm{mm}$ & $\mathrm{CD}/\mathrm{mm}$ & $\mathrm{P}_{1\mathrm{cm}}/\%$ & $\mathrm{R}_{1\mathrm{cm}}/\%$ \\
            \midrule
            First view ($G=576$) & $2.776$ & $2.439$ & $95.56$ & $97.55$ \\
            Best view & $2.586$ & $2.290$ & $96.43$ & $97.96$ \\
            \midrule
            Depth loss only & $2.866$ & $2.918$ & $94.90$ & $93.49$ \\
            SDF loss only & $3.464$ & $2.827$ & $94.58$ & $97.66$ \\
            \midrule
            Init only & $8.095$ & $6.432$ & $67.26$ & $87.75$ \\
            No shape opt. & $2.917$ & $2.582$ & $95.28$ & $97.22$ \\
            Mean shape init & $3.066$ & $2.612$ & $93.75$ & $97.14$ \\
            \midrule
            \hln{Quaternion} & \hln{$3.866$} & \hln{$3.193$} & \hln{$90.67$} & \hln{$96.15$} \\
            \hln{Finer grid ($G=4608$)} & \hln{$3.458$} & \hln{$2.888$} & \hln{$92.64$} & \hln{$96.83$} \\
            \bottomrule
        \end{tabular}
    \end{threeparttable}
\end{table}

\subsection{Run Time Analysis}\label{section:runtime}
In general, analysis-by-synthesis approaches impose run time limitations on the network and components used in the optimization loop. Here, we break down the run time of our pipeline into its different components, namely, the initialization network, the SDF decoder, and the differentiable renderer. Table \ref{tab:computation} shows the resulting run times on a GeForce GTX 1070 Mobile when optimizing for 50 iterations, which is typically sufficient for convergence (rendering is performed at a resolution of $640\times480$). Our custom CUDA implementation achieves rendering times below \SI{6}{ms} (for a single forward and backward pass), which is significantly faster than the decoding step. This indicates that the bottleneck in our analysis-by-synthesis pipeline is the decoding of the latent representation into the SDF, not the differentiable rendering. Including the decoding step, rendering a single image takes around \SI{30}{ms}. 

To compare with the rendering of continuous neural field representations, DIST \cite{liu2020dist}, a differentiable renderer for continuous SDFs, requires \SI{0.99}{\second} to render a single $512\times 512$ image (on a GeForce GTX 1080 Ti) despite various tricks to improve the run time. This highlights the drawback of neural field representations, which require thousands of evaluations during rendering. The discretized shape representation of this work only requires a single forward pass to render from the latent variables.

\begin{table}[tbh]
    \centering
    \begin{threeparttable}
        \caption{Run time breakdown of the pipeline}\label{tab:computation}
        \begin{tabular}{@{}rrr@{}}
            \begin{tabular}{@{}rrr@{}}
                \toprule
                & Time ($\mathrm{ms}$) & \% of Total \\
                \midrule
                Segmentation ($\times1$) & $268.23$ & $15.50$ \\
                Initialization ($\times1$) & $4.67$ & $0.27$ \\
                Decoding-Forward ($\times50$) & $46.22$ & $2.67$ \\
                Rendering-Forward ($\times50$) & $8.53$ & $0.49$ \\
                Losses ($\times50$) & $136.06$ & $7.86$ \\
                Decoding-Backward ($\times50$) & $910.98$ & $52.63$ \\
                Other-Backward ($\times50$) & $247.41$ & $14.29$ \\
                \midrule
                Total & $1731.02$ & \\
                \bottomrule
            \end{tabular}
        \end{tabular}
    \end{threeparttable}
\end{table}

\section{Limitations}

General limitations inherent to analysis-by-synthesis methods apply to our method. Although our method achieves better results than existing discriminative methods, this improvement comes at the cost of slower run time. Furthermore, the complexity of all components included in the optimization loop must be limited to avoid slow optimization. Therefore, we chose to use one VAE per category to limit the network size of the decoder. 

\hln{A possible way towards real-time application would be to use a more complex network for the initialization. This might allow for direct inference of a higher-quality estimate and would reduce the number of necessary optimization iterations. Furthermore, iterative optimization could be performed online with changing sensor data in a tracking fashion \cite{deng2022icaps}.} %

We observed that shape estimation is currently mostly limited to interpolations of shapes seen in training. Generalization to novel shapes is limited. Training on more shapes, cross-category, and other generative models might be possible ways towards more general shape estimation.

Furthermore, our method does not take into account the color information, and pose and shape are estimated based on depth data only.

\section{Conclusion and Outlook}
We presented an architecture to estimate pose and shape of an object. The architecture can be used for single- and multi-view estimation. Our approach is trained on unconstrained orientations and is capable of handling ambiguous views during training due to our $\mathrm{SO}(3)$ parametrization. If the poses in an environment are known to be constrained, such constraints can easily be incorporated into our framework. We open-source our approach to facilitate further research in this area.

Our modular architecture naturally lends itself to various extensions and modifications. The differentiable renderer is currently only used during inference and to generate a depth map. Modifying it to render other modalities and using it for end-to-end training with unannotated data could be an interesting research direction. Other future directions include multi-category models, single-stage training, and a fully probabilistic pose (and shape) estimation framework.

\section*{Acknowledgement}
This work was partially supported by the Wallenberg AI, Autonomous Systems and Software Program (WASP) funded by the Knut and Alice Wallenberg Foundation. 
The authors thank Raghav Bongole for his contributions to the software repository.

\bibliographystyle{IEEEtran}
\bibliography{IEEEabrv,bib}

\end{document}